\documentclass[letterpaper, 10 pt, preprint]{ieeeconf}

\IEEEoverridecommandlockouts                              

\usepackage{array}
\usepackage{textcomp}
\usepackage{stfloats}
\usepackage{cuted}
\usepackage{url}
\usepackage{verbatim}
\def\BibTeX{{\rm B\kern-.05em{\sc i\kern-.025em b}\kern-.08em
    T\kern-.1667em\lower.7ex\hbox{E}\kern-.125emX}}
\usepackage{balance}

\usepackage{amsfonts}
\usepackage{xcolor}
\usepackage{float}
\usepackage{graphicx}
\usepackage{booktabs}
\usepackage{amssymb}
\usepackage{amsmath}
\usepackage{algorithm}
\usepackage{algpseudocode}
\usepackage{adjustbox}
\usepackage{booktabs}
\usepackage{capt-of}
\usepackage{cuted}
\usepackage{textcomp}
\usepackage[table]{xcolor}
\usepackage{pifont}

\title{\LARGE \bf
Flatness Preserves Instruction Following in Vision-Language-Action Models
}

\IEEEaftertitletext{%
\vspace*{-2.0em}
\begin{center}
    \includegraphics[width=0.85\textwidth]{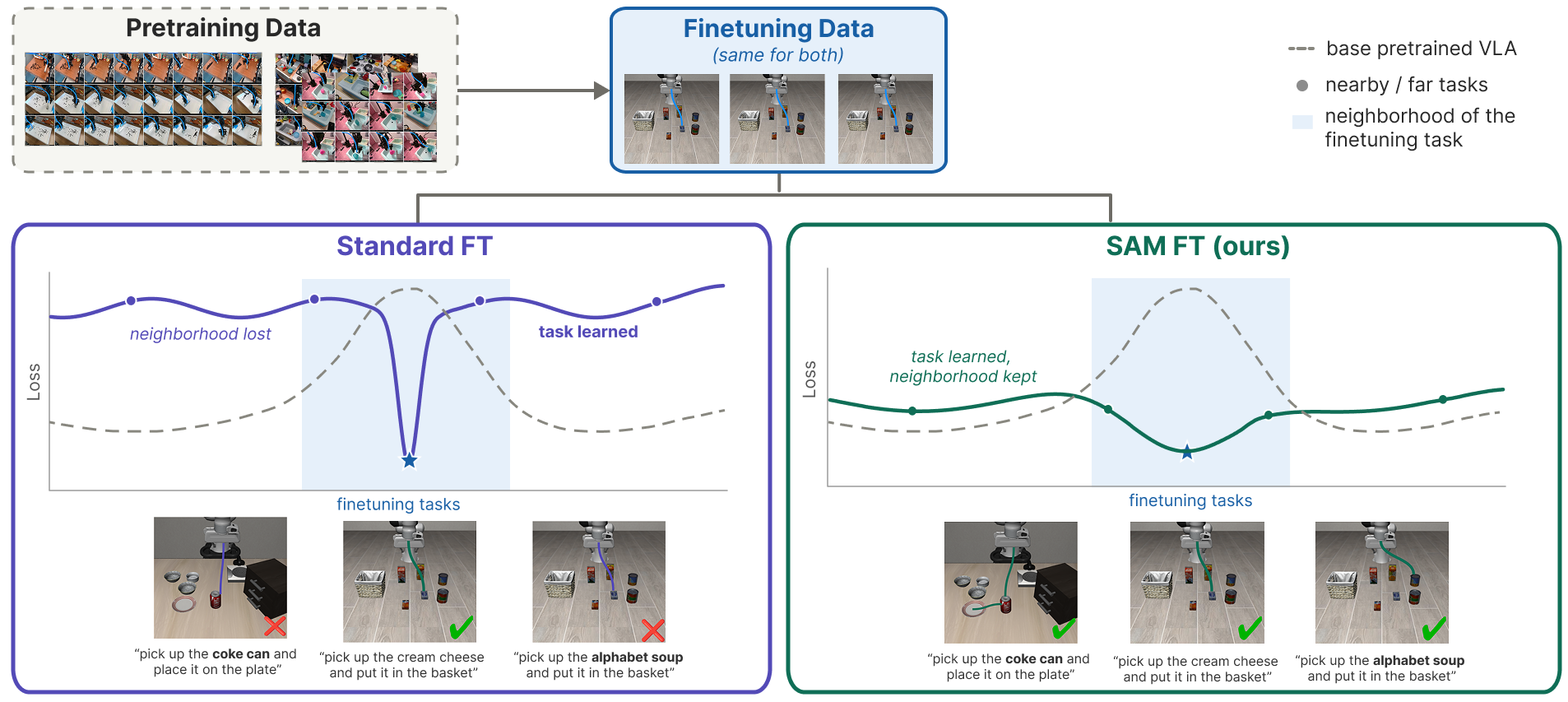}
    \captionof{figure}{Standard finetuning with limited data results in overfitting on seen tasks while forgetting other pretrained knowledge. Finetuning with SAM provides a way to preserve nearby coherence with flatness while finetuned on the same data.}
    \label{fig:overall}
\end{center}
\vspace*{1.0em}
}

\author{Haochen Zhang$^{1}$ \and 
Yonatan Bisk$^{1}$ 
\thanks{$^{1}$The authors are with Carnegie Mellon University, Robotics Institute, Pittsburgh, PA.
{\tt\small\{haochen4, ybisk\}@andrew.cmu.edu}}
}

\begin{document}
\maketitle
\thispagestyle{empty}
\pagestyle{empty}
\vspace*{-3em}


\vspace*{0.8em}
\begin{abstract}
Vision-language-action (VLA) models have the potential for open-world generalization by leveraging pretrained vision-language representations, yet downstream finetuning on limited robot data often degrades these representations. This leads to brittle action policies that ignore language instructions in favor of visual shortcuts, despite the base VLM having this knowledge, which is a failure mode we term \textit{instruction blindness}. We hypothesize that standard finetuning with limited data applies gradients to a sparse set of points, which manifests as a sharp loss landscape with high-curvature minima. We analyze this effect, then show that regularization through explicit flatness-preserving optimization is a simple yet effective technique while finetuning on the exact same data.
Specifically, we demonstrate that applying sharpness-aware minimization (SAM) during VLA finetuning significantly improves instruction following by over 55\% across multiple simulation and real-world benchmarks without additional data, architectural modification, or pretraining. Explicitly preserving flatness better preserves pretrained knowledge compared to other regularization techniques that simply anchor model weights or constrain learning. We further analyze the effect of selective sharpness, quantify its effects, and show that such approaches are complementary to existing guidance techniques.
\end{abstract}

\section{Introduction}

There is debate in the robotics community as to how vision-language-action (VLA) models should leverage pretrained vision-language model (VLM) backbones \cite{barreiros2026careful, zhang2026vlm4vla, grover2025enhancing}. For VLMs, internet-scale pretraining provides good coverage of the latent space for subsequent test samples. Leveraging these representations in VLA models provides the potential for robot policies to possess open-world knowledge only if they can be transferred to robot action. However, numerous works have shown the degradation of pretrained representations and brittleness of VLA models to small input perturbations, especially when fine-tuned on low-data regimes \cite{grover2025enhancing, guo2025robustness, zhou2025libero, fang2026vision, xu2025seeing, huang2026breaking}. 
One significant failure mode is \textit{instruction blindness} \cite{zhou2025libero, zhan2026stable}, where the model disregards the language instruction due to vision shortcuts learned during finetuning. Different from other instruction following failures, instruction blindness is not due to out-of-distribution language with respect to the VLM backbone, but the degradation of pretrained representations after the finetuning process.
For example, given a scene trained with the instruction ``pick up the cream cheese'', the model will disregard new instructions such as ``pick up the butter'' and blindly carry out the training task \cite{zhang2026vlm4vla, zhan2026stable, xu2025seeing, lian2026bayesianvla}, despite ``butter'' being within the pretraining data. To mitigate this, existing methods often use new pre-training methods with data augmentation \cite{grover2025enhancing, yang2025instructvla, lian2026bayesianvla, hancock2025actions}, guidance techniques \cite{fang2026vision, zhang2026restoring}, or combinations of these \cite{zhan2026stable}. We hypothesize that while such techniques are valid, they may simply be trying to remedy a more fundamental issue with regards to representation loss that can be directly addressed.

During VLM pretraining, internet-scale data provides dense coverage of the vision-language space such that for any new test sample, there is high probability that a nearby embedding was seen during training. LLMs and VLMs are trained with techniques to enforce such neighborhood coherence. Prior to deployment, VLA models are usually finetuned on a significantly smaller in-domain robot dataset, where data imbalance is common.
Gradients are applied only to a sparse set of demonstration samples, which easily causes overfitting. At the embedding level, smoothness is lost as the vast majority of the embedding space, including neighboring regions, have no guarantee of mapping to nearby representations. Correspondingly at the parameter level, we hypothesize that over-indexing on few samples produces a \textit{sharp loss landscape} where the model converges to high-curvature minima and resulting policies are highly brittle to new inputs.

Intuitively, a sharper model is more sensitive to parameter perturbations and thus will be sensitive to variations in the input data as well. Recent work has also shown that models sensitive to parameter updates exhibit ``forgetting'' of their pretraining abilities \cite{watts2026sharpness}. In the machine learning (ML) and natural language processing (NLP) communities, \textit{flatness-preserving optimization methods} are widely used as a remedy to improve generalization and reduce catastrophic forgetting \cite{watts2026sharpness, foret2020sharpness, kwon2021asam, sherborne2023tram, liu2022towards}.
In particular, sharpness-aware minimization (SAM) \cite{foret2020sharpness} is a form of bilevel optimization that enforces the model towards weight spaces in uniformly flatter regions by minimizing the worst-case loss within a region around the current parameters at each step. By enforcing flatter loss basins, the learned embedding function cannot be too sensitive to similar inputs in the neighborhood of seen data, since both would manifest as loss instability. We test the effect of SAM on VLA models and what this uncovers about the representations in such models.
 
Through both simulation experiments across three counterfactual instruction benchmarks and real-world experiments, \textbf{we show that by applying SAM during VLA finetuning on the exact same data, the robustness and instruction-following ability of the VLA model dramatically improves on counterfactual instructions} by 60.2\%, 55.9\%, and 217\% on the LIBERO-PRO Task, LangGap, and LIBERO-CF benchmarks respectively. While other standard regularization methods can help maintain pretrained representations, SAM in particular explicitly enforces flatter minima and is consistently more effective at preserving language-following while maintaining train accuracy. Additionally, we argue that this method does not render other approaches obsolete, but instead provides a substantially higher baseline for the community to leverage in a complementary way. Importantly, we show that performance improves \textbf{without requiring more data coverage during finetuning}. We then explore component-based application of SAM, quantify the sharpness, and analyze the effects of SAM optimization used in combination with other techniques. To the best of our knowledge, this is the first work to explore flatness-preservation techniques to robot policies.
 
\section{Related Work}

\subsection{Preserving Pretrained Representations in VLAs.}

\begin{table}[t]
\centering
\small
{\setlength{\tabcolsep}{1pt}
\renewcommand{\arraystretch}{1.0}
\caption{Comparison of standard regularization methods.}
\label{tab:regularization_comparison}
\begin{tabular}{@{}l@{\hspace{-3pt}}c@{\hspace{4pt}}c@{}}
\toprule
\textbf{Method} & \textbf{Backbone Changed?} & \textbf{What Is Preserved} \\
\midrule
Lower LR & \checkmark & Smaller shifts in weights \\
LoRA & \ding{55} & Pretrained weights \\
Weight Decay & \checkmark & Small parameter magnitudes \\
EMA/SWA & \checkmark & Averaged model weights \\
SAM & \checkmark & Flatness of the loss landscape \\
\bottomrule
\end{tabular}}
\vspace{-1.5em}
\end{table}

Existing works retain pretrained representations with various data, architecture, training, or inference techniques. Some methods \cite{grover2025enhancing, yang2025instructvla, hancock2025actions} leverage additional co-training data, new annotations, and architectural changes. This can be costly in terms of data collection, full re-training, or complex architectural changes.
Methods that regularize the loss or weight space are most similar to our approach. There are already a number of standard ways to regularize a model, and different methods vary in how they allow model weights to change over training. One of the simplest methods is to lower the learning rate, which has been found to improve flatness \cite{rofin2026learning} by taking smaller effective gradient steps. Other standard regularization methods include low-rank adaptation (LORA) \cite{hu2022lora}, weight decay, stochastic-weight averaging (SWA) \cite{izmailov2018averaging}, exponential moving average (EMA), parameter merging \cite{yadav2025robust}, and sharpness aware minimization (SAM) \cite{foret2020sharpness}. 
A functional comparison of these is in Table \ref{tab:regularization_comparison}.

\subsection{Instruction Faithfulness.}
Recently, ensuring that policies faithfully follow language instructions has emerged as a distinct focus. Data-driven approaches \cite{glossop2025cast, hou2026langgap} address this through targeted finetuning dataset construction while representation-level methods \cite{zhang2026restoring} up-weight language signals. Similarly, other work \cite{lian2026bayesianvla, xu2025seeing, fang2026vision} decouples the visual prior and applies guidance at inference time to steer the policy. 
These approaches, however, are not as necessary if the pretrained representations are preserved in the first place, which we directly target loss landscape geometry to achieve.
Most similar to our approach, a concurrent work \cite{huang2026breaking} focuses on directly regularizing the weight space during VLA fine-tuning in combination with inference-time guidance. Our approach is a distinct alternative, which provides a higher performance baseline that can be used in combination with other techniques. 

\subsection{Optimization of Loss Landscape.}
The relationship between loss landscape geometry and generalization has been extensively studied in ML literature \cite{hochreiter1997flat, keskar2016large, jiang2019fantastic, foret2020sharpness, sherborne2023tram, Bahri2022sharpness,Yang2025sharpzo}. \cite{hochreiter1997flat} first formalized the connection between flat minima and generalization, observing that solutions residing in wide regions of the loss surface tend to generalize better than those lying in sharp, narrow ones. Methods like stochastic weight averaging over model weights can implicitly achieve flatness while sharpness-aware minimization (SAM) \cite{foret2020sharpness} is an explicit bi-level optimization which has been shown to improve generalization across vision, language, and joint tasks \cite{Bahri2022sharpness, liu2022towards, sherborne2023tram, Yang2025sharpzo}. Subsequent work has proposed a number of variants such as ASAM \cite{kwon2021asam}, TRAM  \cite{sherborne2023tram}, and F-SAM \cite{li2024friendly} to reduce parameter tuning, improving generalization, or boost efficiency.
\section{Problem Formulation}
\subsection{Preliminaries}
\textbf{Vision Language Action Models.}
VLA models typically consist of a VLM backbone and a flow-matching or diffusion action head conditioned on vision and language inputs. These models are a near perfect candidate for this problem as they are sequentially trained extensions of an initial pretrained knowledge backbone. A VLA model is denoted as \(\pi(a | o,l)\). \(a\) is the \(k\)-step future action chunk, \(o\) is the current visual observation (e.g. RGB wrist camera and external camera images), and \(l\) is the given language instruction. We assume access to robot finetuning data, \(D_{FT}\), consisting of \((a, o, l)\) tuples, and \(D_{PT} \gg D_{FT}\) for internet-scale pretraining data \(D_{PT}\). The supervised finetuning (SFT) objective is typically:
\begin{equation}
    \mathcal{L}(\theta) = \mathbb{E}_{t, x_0, x_1}\left[\| v_\theta(x_t, t, o, l) - (x_1 - x_0) \|^2\right]
\end{equation}
where a learned velocity field $v_\theta(x_t, t, o, l)$ at time \(t\) transports Gaussian noise $x_0 \sim \mathcal{N}(0,I)$ to the target action distribution $x_1 = a$ along a linear path $x_t = (1-t)x_0 + tx_1$. 

\textbf{Instruction Blindness.}
A key failure mode arising from vision bias is \textit{instruction blindness}.
Given a finetuning dataset \(D_{FT}\) with samples \((o,l)\) consisting of an observation $o$ and language instruction $l$, a test sample \((o,l')\)
is \textit{out-of-combination} (OOC) if the pair was not observed during training, even if \(o\) and \(l'\) were seen independently. Our first assumption here is that \((o,l')\) test pairs still remain \textit{within the support} of the pretrained action distribution of the base policy (i.e. grasps, joint angles, and positions). More formally, $a \in \text{supp}(\pi_\theta(a))$ for all test tasks.
Instruction blindness is then a practically significant case of OOC, in which the VLA model does not follow the language instruction: 
\begin{equation}
\pi(a | o,l') \simeq \pi(a | o)
\end{equation}
Due to the small set of finetuned trajectories, the behavior exhibits a previously seen training task:
\begin{equation}
\pi(a | o,l') \simeq \pi(a | o,l), l' \neq l
\end{equation}
where \(l\) is a previously observed instruction for the same scene but is semantically distinct from \(l'\).
From a geometric perspective, instruction blindness occurs when the learned manifold \(\mathcal{M}\) is insufficiently sensitive to semantically distinct language: the manifold has converged such that novel \((o,l)\), which are within support of the internet-scale pretrained VLM, are now poorly supported on \(\mathcal{M}\). Instruction blindness has been shown to be a failure case of VLAs in numerous existing works \cite{grover2025enhancing, fang2026vision, xu2025seeing, huang2026breaking, lian2026bayesianvla, zhan2026stable}. 

\textbf{Flatness and Generalization.}
The existence of \textit{sharp minima} is common in large-scale, overparameterized models, where multiple types of minima can yield similar training loss while having significantly different test performance \cite{sherborne2023tram}. Formally, \cite{foret2020sharpness} defines the sharpness \(\phi\), of a solution \(\theta\), as the worst-case loss perturbation in a neighborhood of radius \(\rho\):
\begin{equation}
    \phi_\rho(\theta) = max_{||\epsilon|| \leq \rho} \mathcal{L}(\theta + \epsilon) - \mathcal{L}(\theta)
    \label{eqn:sharpness}
\end{equation}
A solution is considered flat \textit{iff} \(\phi_\rho(\theta)\) remains small for non-trivial \(\rho\), indicating that the learned model is robust to parameter perturbations. A large body of work has linked sharp minima to weaker generalization, increased catastrophic forgetting, and degraded alignment in models such as LLMs and VLMs \cite{foret2020sharpness, sherborne2023tram, watts2026sharpness, chen2026unveiling, zhang2026lost}. \cite{foret2020sharpness} specifically showed that flat minima admit tighter PAC-Bayes generalization bounds, and thus solutions stable under parameter-space perturbations result in better generalization to test samples. We hypothesize that this is suitable in the low-data VLA finetuning setting as well, where model parameters need to shift to learn to output actions in the downstream domain, but simultaneously maintain some of the representational structure from pretraining. For the manifold \(\mathcal{M}\) learned by the VLA, we seek a \textit{smooth manifold} with curvature that varies gradually, such that nearby regions shift together as the model is finetuned. 
Our second assumption is that the capability of grounding the OOC language instructions still exists within the original VLM backbone. That is, the pretrained representations encode sufficient information to distinguish between language instructions, even if this signal is suppressed during downstream finetuning.

\subsection{Setup}
Following \cite{foret2020sharpness, keskar2016large}, we quantify sharpness as a normalized version of \(\phi\):
\begin{equation}
\label{eqn:keskar}
    \mathcal{S}(\theta) = \frac{\phi_{\rho}(\theta)}{1 + 
    \mathcal{L}(\theta)} \times 100
\end{equation}
\vspace{-1pt}
We adopt the normalization to make sharpness values comparable across model scales, or training stages when the baseline loss itself varies substantially in absolute value. We maintain the $\rho$-ball for calculating worst case loss to keep the calculation tractable at our parameter scale.

Let $\mathcal{T}(o, l)$ denote a task with visual observation $o$ and language instruction $l$, and $\mathbf{1}[\pi_\theta, \mathcal{T}]$ 
be an indicator of task success under policy $\pi_\theta$. We define the 
instruction following performance as:
\begin{equation}
    \mathcal{F}(\theta) = \mathbb{E}_{(o,l) \notin D_{FT}} 
    \left[\mathbf{1}[\pi_\theta, \mathcal{T}(o,l)]\right]
\end{equation}

\vspace{-2pt}

Given \(D_{FT}\) with vision-language imbalance and lower language support, we seek to find parameters \(\theta^*\) such that: \(\mathcal{S}(\theta^*) < \mathcal{S}(\theta)\),
where $\theta$ denotes the parameters obtained via standard finetuning. We hypothesize that this reduction in sharpness is associated with 
improved instruction following on OOC pairs: \(\mathcal{S}(\theta^*) < \mathcal{S}(\theta) \implies 
    \mathcal{F}(\theta^*) > \mathcal{F}(\theta)\),
which we evaluate empirically in Section~\ref{sec:result}.
\section{Method}

\begin{algorithm}[h]
\caption{SAM Finetuning for VLAs}
\label{alg:sam_vla}
\begin{algorithmic}[1]
\footnotesize
\Require Pretrained VLA $\pi_\theta$, finetuning dataset $D_{FT}$, 
         perturbation radius $\rho$, learning rate $\eta$, 
         base optimizer AdamW
\Ensure Finetuned parameters $\theta^*$
\While{not converged}
    \State Sample batch $(a, v, l) \sim D_{FT}$
    \State \textbf{// Step 1: Compute perturbation at clean parameters}
    \State $\mathcal{L}(\theta) \leftarrow \mathbb{E}_{t, x_0, x_1}\left[\| v_\theta(x_t, t, v, l) - (x_1 - x_0) \|^2\right]$
    \State $g \leftarrow \nabla_\theta \mathcal{L}(\theta)$
    \State $\hat{\epsilon} \leftarrow \rho \cdot g \;/\; \|g\|$
    \State \textbf{// Step 2: Compute gradient at perturbed parameters}
    \State $\tilde{g} \leftarrow \nabla_\theta 
           \mathcal{L}(\theta + \hat{\epsilon})$
    \State \textbf{// Step 3: Update original parameters with AdamW}
    \State $\theta \leftarrow \text{AdamW}(\theta,\; \tilde{g},\; \eta)$
\EndWhile
\State \Return $\theta^*$
\end{algorithmic}
\end{algorithm}

We aim to mitigate instruction blindness and improve generalization in VLA finetuning without additional data collection, pretraining, or architectural modifications by applying sharpness-aware minimization (SAM) \cite{foret2020sharpness} to VLA finetuning.
SAM enforces a notion of flatness in the loss landscape to learn model parameters lying in a flatter loss basin, making them more stable against parameter perturbations. Prior work, however, has primarily shown the benefits of SAM on LLMs and vision models during finetuning. In the robotics domain, the utility of SAM on a multi-stage trained model trained from a VLM backbone, to a base VLA, to a finetuned VLA, is unique and worth investigating. Our intuition is that explicitly limiting sensitivity to local perturbations and considering the surrounding loss neighborhood will allow the model to preserve some of the neighborhood coherence in the VLM backbone, which is crucial for limited-data finetuning with robotics data.
We verify this hypothesis empirically through extensive experiments in Section~\ref{sec:result}. 

Specifically, SAM \cite{foret2020sharpness} reformulates standard gradient descent as a bi-level optimization problem that penalizes sharp minima. 
Rather than minimizing $\mathcal{L}(\theta)$ directly, SAM seeks parameters \(\theta^*\) that 
minimize the worst-case loss in a $\rho$-neighborhood of $\theta$, where \(\rho\) is tuned as a hyperparameter:
\begin{equation}
    \theta^* = \arg\min_\theta \max_{\|\epsilon\| \leq \rho} 
\mathcal{L}(\theta + \epsilon)
\end{equation}

In practice, this is approximated via a two-step gradient computation at each training iteration. First, the maximizing perturbation is estimated via a single gradient ascent step:
\begin{equation}
\hat{\epsilon}(\theta) = \rho \cdot \frac{\nabla_\theta 
\mathcal{L}(\theta)}{\|\nabla_\theta \mathcal{L}(\theta)\|}
\end{equation}
The model parameters are then updated using the gradient evaluated at the 
perturbed point $\theta + \hat{\epsilon}(\theta)$, rather than at $\theta$:
\begin{equation}
\theta \leftarrow \theta - \eta \nabla_\theta 
\mathcal{L}(\theta + \hat{\epsilon}(\theta))
\end{equation}
where $\eta$ is the learning rate.
Algorithm \ref{alg:sam_vla} shows the pseudocode for the SAM algorithm on a VLA.

\section{Evaluation in Simulation}
\label{sec:result}

Broadly, we are interested in understanding: \textbf{RQ 1}) Do the generalization benefits of SAM transfer to VLA models, specifically by mitigating instruction blindness?
\textbf{RQ 2}) What components or representations does SAM finetuning change?
\textbf{RQ 3}) Which part of the VLA model has the highest sharpness and how should SAM be applied?

We finetune a pre-trained VLA model using SAM and benchmark language-following ability compared with a variety of baselines finetuned on the exact same (or more) data. The baselines are chosen to represent different paradigms for preserving pretrained representations or improving adherence to language. We evaluate the methods on three counterfactual benchmarks: LIBERO-PRO Task \cite{zhou2025libero}, LIBERO-CF \cite{fang2026vision}, and LangGap \cite{hou2026langgap}, totaling 138 tasks. These benchmarks evaluate OOC tasks on training scenes, allowing us to evaluate instruction blindness.
We obtain answers to our research questions as follows, with the details in Sections C-E:
\begin{itemize}
    \item \textbf{RQ1}: Yes, applying SAM during finetuning significantly improves instruction following.
    \item \textbf{RQ2}: Quantitatively, sharpness in the loss landscape is reduced and coherence is verified qualitatively with a simple representation probe.
    \item \textbf{RQ3}: The highest sharpness is in the action head though global application of SAM works best.
\end{itemize}

\subsection{Baseline Models}
\textbf{Vanilla VLAs.}
We first evaluate off-the-shelf VLA models finetuned on LIBERO, including OpenVLA-OFT \cite{kim2025fine}, and $\pi_{0.5}$ \cite{black2025pi}. OpenVLA-OFT uses a Prismatic 7B VLM \cite{karamcheti2024prismatic} to decode action tokens. $\pi_{0.5}$ uses a PaliGemma \cite{steiner2024paligemma} backbone with a flow-matching action head\footnote{Note: ``knowledge insulation'' techniques \cite{driess2026knowledge} and EMA were already used in the pretrained $\pi_{0.5}$ checkpoint which we further finetune on.}. These models represent the effect of standard finetuning practices on biased data. 

\textbf{Bayesian-Factorized Models.}
In recent literature, methods explicitly factorize the VLA model into a prior \(\pi(a | o)\) and likelihood model \(\pi(l|o, a)\) which are then combined via the optimization objective \cite{lian2026bayesianvla} or with guidance at inference time \cite{fang2026vision, zhan2026stable}. We compare with the
CAG \cite{fang2026vision} model, and our own classifier-free guidance \cite{ho2022classifier} variant of the default VLA, denoted as $\pi_{0.5}$\_cfg, finetuned with language dropout.\footnote{BayesVLA only compares on the LIBERO-Object subset and we are unable to reproduce their baseline.}

\textbf{Data-Augmented Models.}
A data-centric approach to preserving pretrained language representations is to pretrain or finetune with more diverse language data, including counterfactual examples.
We compare with a representative checkpoint from this paradigm \cite{hou2026langgap}, denoted as \(\pi_{0.5}\)\_LangGap, where the model is finetuned with a diverse set of 45 tasks and counterfactual language.

\textbf{Models with Representation Regularization.}
As our goal is to avoid forgetting knowledge in the backbone, we compare standard regularization methods with SAM.  
We evaluate baselines with each of these techniques and compare to SAM to understand the effect of generic regularization techniques compared to ones that explicitly preserve flatness.

\subsection{Implementation and Metrics}
We apply SAM to $\pi_{0.5}$\_base during LIBERO finetuning with $\rho$ tuned as a hyperparameter. The base optimizer is AdamW \cite{loshchilov2017decoupled} with gradient clipping. Each finetuned model was finetuned with 30k steps on the same LIBERO training data with 40 tasks. All evaluations are run zero-shot on new \((o, l)\) pairs with 50 rollouts. We also apply SAM to OpenVLA-OFT and verify that instruction following improves on a different architecture, but focus on the SOTA \(\pi_{0.5}\) model as it provides a higher performance baseline than OpenVLA-OFT.

We primarily evaluate instruction-following ability by reporting the task success rate (SR). To quantify the sharpness of the learned loss landscape, we use a normalized sharpness (Eqn. \ref{eqn:keskar}), \(S\),
and the eigenvalue spectrum of the Hessian $\nabla^2_\theta \mathcal{L}(\theta)$, where the largest eigenvalue $\lambda_{\max}$, provides a local measure of curvature at the solution. We use a Lanczos approximation to calculate this, following \cite{foret2020sharpness}.
For real-world experiments, we also measure the Grounding Rate defined in \cite{fang2026vision} as a binary score per rollout indicating whether the gripper makes contact with the target object. This effectively measures the vision-language grounding ability of the model through it's action ``intent'' without penalizing grasp failures. We note this is an important metric for real-world experiments as full completion of an action trajectory is more challenging, while instruction blindness concerns the ability to ground the natural language instruction correctly.

\subsection{RQ 1: Instruction Following}
\label{RQ1}
\begin{figure*}[t]
    \centering
    \includegraphics[width=0.86\linewidth]{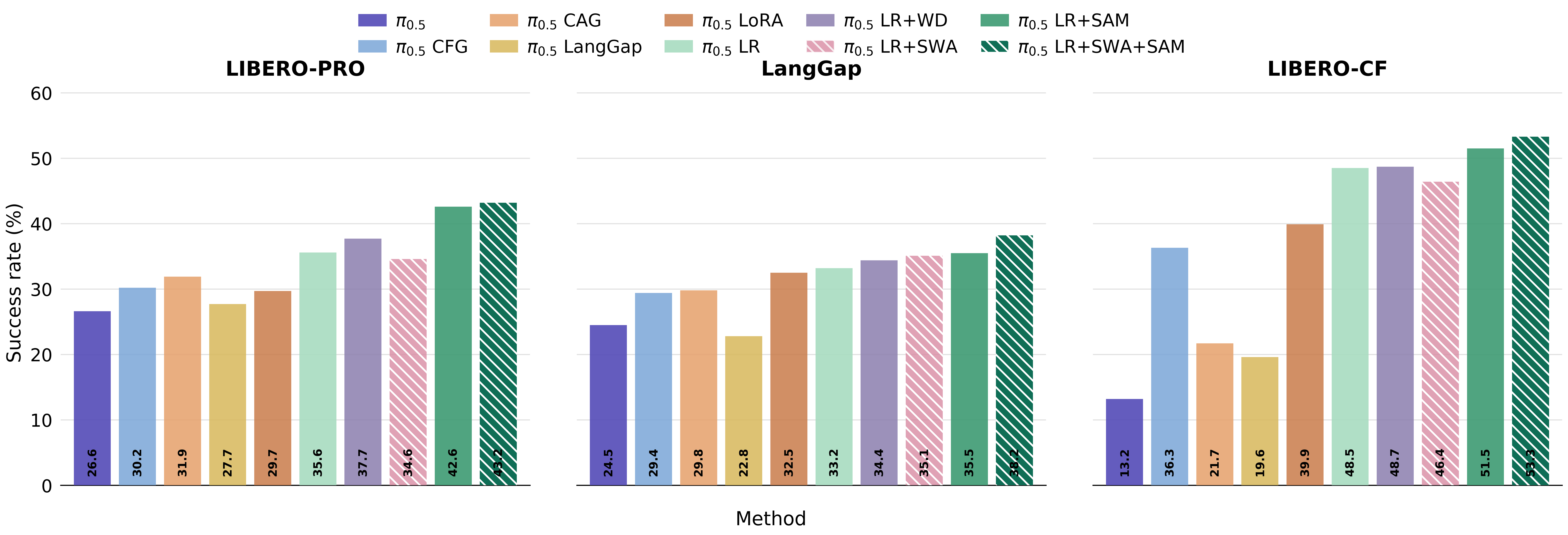}
    \caption{Comparison across baselines on three counterfactual instruction-following benchmarks}
    \label{fig:results_chart}
    \vspace{-1em}
\end{figure*}

\begin{table*}[h!]
\centering
\footnotesize
\caption{Success rate (\%) of representation-regularization methods on LIBERO-PRO \cite{zhou2025libero}, LIBERO-LangGap \cite{hou2026langgap}, and LIBERO-CF. All variants use the same learning rate schedule and 50 rollouts per task.}
\label{tab:regularization_splits}
{\setlength{\tabcolsep}{2.7pt}
\renewcommand{\arraystretch}{1.05}
\begin{adjustbox}{max width=\textwidth}
\begin{tabular}{@{}l*{5}{c}@{\hspace{4pt}\vrule\hspace{4pt}}*{4}{c}@{\hspace{4pt}\vrule\hspace{4pt}}*{5}{c}@{}}
\toprule
& \multicolumn{5}{c@{\hspace{4pt}\vrule\hspace{4pt}}}{\textbf{LIBERO-PRO}}
& \multicolumn{4}{c@{\hspace{4pt}\vrule\hspace{4pt}}}{\textbf{LangGap}}
& \multicolumn{5}{c}{\textbf{LIBERO-CF}} \\
\cmidrule{2-6} \cmidrule{7-10} \cmidrule{11-15}
\textbf{Method}
& \textbf{Goal} & \textbf{Spatial} & \textbf{Object} & \textbf{Long} & \textbf{Avg}
& \textbf{Goal} & \textbf{Spatial} & \textbf{Object} & \textbf{Avg}
& \textbf{Spatial} & \textbf{Object} & \textbf{Long} & \textbf{OOD} & \textbf{Avg} \\
\midrule
$\pi_{0.5}$ LoRA {\scriptsize (S=0.032, $\lambda_{\max}$=9.04)}
& 34.0 & 28.8 & 47.6 & 8.4 & 29.7
& 47.1 & 17.8 & 45.3 & 32.5
& 24.0 & \textbf{43.4} & 58.2 & 34.0 & 39.9 \\

$\pi_{0.5}$ LR {\scriptsize (S=0.022, $\lambda_{\max}$=9.04)}
& 38.4 & 51.6 & 37.8 & 14.6 & 35.6
& \textbf{52.0} & 16.9 & 46.1 & 33.2
& 50.7 & 35.6 & 53.2 & 54.3 & 48.5 \\

$\pi_{0.5}$ LR+WD {\scriptsize (S=0.021, $\lambda_{\max}$=7.70)}
& 38.0 & 59.4 & 34.0 & 19.4 & 37.7
& \textbf{52.0} & 18.1 & 48.0 & 34.4
& 52.0 & 37.4 & 52.8 & 52.4 & 48.7 \\

$\pi_{0.5}$ LR+SWA {\scriptsize (S=0.015, $\lambda_{\max}$=7.52)}
& 25.4 & 63.4 & 36.0 & 13.4 & 34.6
& 45.3 & 22.2 & 47.3 & 35.1
& \textbf{54.5} & 33.6 & 51.0 & 46.7 & 46.4 \\

$\pi_{0.5}$ LR+SAM {\scriptsize (S=0.010, $\lambda_{\max}$=3.91)}
& \textbf{42.4} & \textbf{54.4} & 47.0 & 24.8 & 42.2
& 49.1 & 19.3 & 50.5 & 35.5
& 53.6 & 41.6 & 60.4 & 50.1 & 51.4 \\

$\pi_{0.5}$ LR+SWA+SAM {\scriptsize (S=0.012, $\lambda_{\max}$=3.98)}
& 37.0 & 53.2 & \textbf{56.0} & \textbf{28.4} & \textbf{43.6}
& 48.7 & \textbf{22.9} & \textbf{53.5} & \textbf{38.2}
& 51.9 & 41.0 & \textbf{61.8} & \textbf{58.3} & \textbf{53.3} \\

\bottomrule
\end{tabular}
\end{adjustbox}}
\vspace{-1em}
\end{table*}

As shown in Fig. \ref{fig:results_chart}, applying SAM during finetuning (solid green bar) with the exact same data consistently outperforms the default VLA (purple bar), achieving SOTA performance across the LIBERO-PRO Task, LangGap, and LIBERO-CF benchmarks by 16\%, 13.7\%, and 28.7\% percentage points respectively. This method also outperforms all relevant baselines across different paradigms, even methods that use additional training data \cite{hou2026langgap}. We note that the Bayesian-Factorization class of models and guidance techniques at inference-time also provide better instruction following compared to the default VLA, but still perform worse than SAM. 

As the main goal of this work is knowledge preservation, the most relevant comparisons are other regularization-based methods.
We verify that simply finetuning the VLA model with a smaller learning rate schedule improves instruction following, aligned with prior LLM work that shows smaller learning rates during finetuning prevent forgetting \cite{rofin2026learning}. 
We then test LoRA, which adapts pretrained weights while keeping most frozen, but performance improvements are not consistent across tasks. The most effective techniques are SWA and weight decay, which constrain model parameters during finetuning. These methods add a small gain on top of the smaller learning rate checkpoint but not as much as SAM. We hypothesize this is because they functionally regularize by anchoring model weights over time, sometimes inducing a flatter loss landscape but this is not guaranteed. SAM aims to preserve pretrained representations by explicitly enforcing flatness while still allowing model weights to shift, achieving the highest overall accuracy on counterfactual instructions. Furthermore, stacking SAM with SWA achieves a further performance boost, especially on LangGap and LIBERO-CF.
Empirically, this strengthens the relation between flatness and vision-language-action generalization.

Next, we examine in Table \ref{tab:regularization_splits} the task subset performance breakdown amongst the strongest regularization-based baselines. We observe that on LIBERO-PRO Task and LangGap, the ``Spatial'' subset achieves significant gain, which involves correctly grounding a specific object by position among similar distractor objects. This suggests that controlling the parameter shift during downstream finetuning helps retain the vision-language coupling to minimize spurious correlations between language and spatial scene layouts. SAM also achieves a large gain on the ``Long'' subset, which has longer-horizon tasks.

\begin{table}[h]
\centering
\scriptsize
\caption{Success rate (\%) for $\pi_{0.5}$ on LIBERO training data and the relative drop in success rate for regularization methods. $^\dagger$ indicates result reported in the paper.}
\label{tab:train_results}
\begin{tabular}{lccccc}
\toprule
\textbf{Method} & \textbf{Goal} & \textbf{Spatial} & \textbf{Object} & \textbf{Long} & \textbf{Average} \\
\midrule
$\pi_{0.5}$$^\dagger$ \cite{black2025pi}
& 98.0 & 98.8 & 98.2 & 92.4 & 96.9 \\
\midrule
$\pi_{0.5}$\_cfg
& \cellcolor{red!10}-2.0
& \cellcolor{red!25}-7.4
& \cellcolor{red!25}-7.6
& \cellcolor{red!35}-13.2
& \cellcolor{red!25}-7.6 \\

$\pi_{0.5}$\_CAG
& \cellcolor{red!15}-5.0
& \cellcolor{red!15}-4.0
& \cellcolor{red!12}-3.6
& \cellcolor{red!30}-10.4
& \cellcolor{red!15}-5.8 \\

$\pi_{0.5}$\_LangGap$^\dagger$ \cite{hou2026langgap}
& \cellcolor{red!35}-13.0
& \cellcolor{red!13}-3.8
& +1.8
& \cellcolor{red!35}-14.4
& \cellcolor{red!20}-7.4 \\
\midrule

$\pi_{0.5}$\_LORA
& \cellcolor{red!20}-6.0
& \cellcolor{red!20}-5.6
& \cellcolor{red!5}-0.6
& \cellcolor{red!45}-19.8
& \cellcolor{red!25}-8.1 \\

$\pi_{0.5}$\_LR
& \cellcolor{red!15}-3.8
& \cellcolor{red!10}-1.8
& +1.0
& \cellcolor{red!8}-1.0
& \cellcolor{red!9}-1.4 \\

$\pi_{0.5}$ LR+WD
& \cellcolor{red!13}-3.2
& \cellcolor{red!13}-2
& \cellcolor{red!5}-0.8
& +0.6
& \cellcolor{red!8}-1.4 \\

$\pi_{0.5}$ LR+SWA
& \cellcolor{red!18}-5.0
& \cellcolor{red!12}-2.6
& \cellcolor{gray!5}0.0
& +0.4
& \cellcolor{red!10}-1.8 \\

$\pi_{0.5}$ LR+SAM
& \cellcolor{red!8}-2.8
& \cellcolor{red!5}-1.4
& \cellcolor{gray!5}0.0
& \cellcolor{gray!5}0.0
& \cellcolor{red!5}-1.1 \\

$\pi_{0.5}$ LR+SAM+SWA
& \cellcolor{red!10}-2.0
& \cellcolor{red!14}-3.4
& \cellcolor{gray!5}0.0
& +0.4
& \cellcolor{red!8}-1.3 \\
\bottomrule
\end{tabular}
\vspace{-0.8em}
\end{table}

Furthermore, we can examine if methods that improve on counterfactual instructions still retain their high performance on training tasks. The success rates of each baseline on the LIBERO training tasks is shown in Table \ref{tab:train_results}. For the guidance-based methods and LoRA, which showed improvements for counterfactual instructions trades this off with performing worse on the original training tasks.
SWA and SAM are the two methods that still maintain performance on training data on par with the default \(\pi_{0.5}\) model. Overall, SAM is a stackable technique and does not preclude other methods for improving instruction following. For example, using a SAM-finetuned model as the conditional model in a CFG framework could further mitigate instruction blindness by providing a higher baseline model to improve upon.

\subsection{RQ 2: Verification of Representation Shifts}
\label{RQ2}

\begin{figure}[h]
    \centering
    \includegraphics[width=0.4\textwidth]{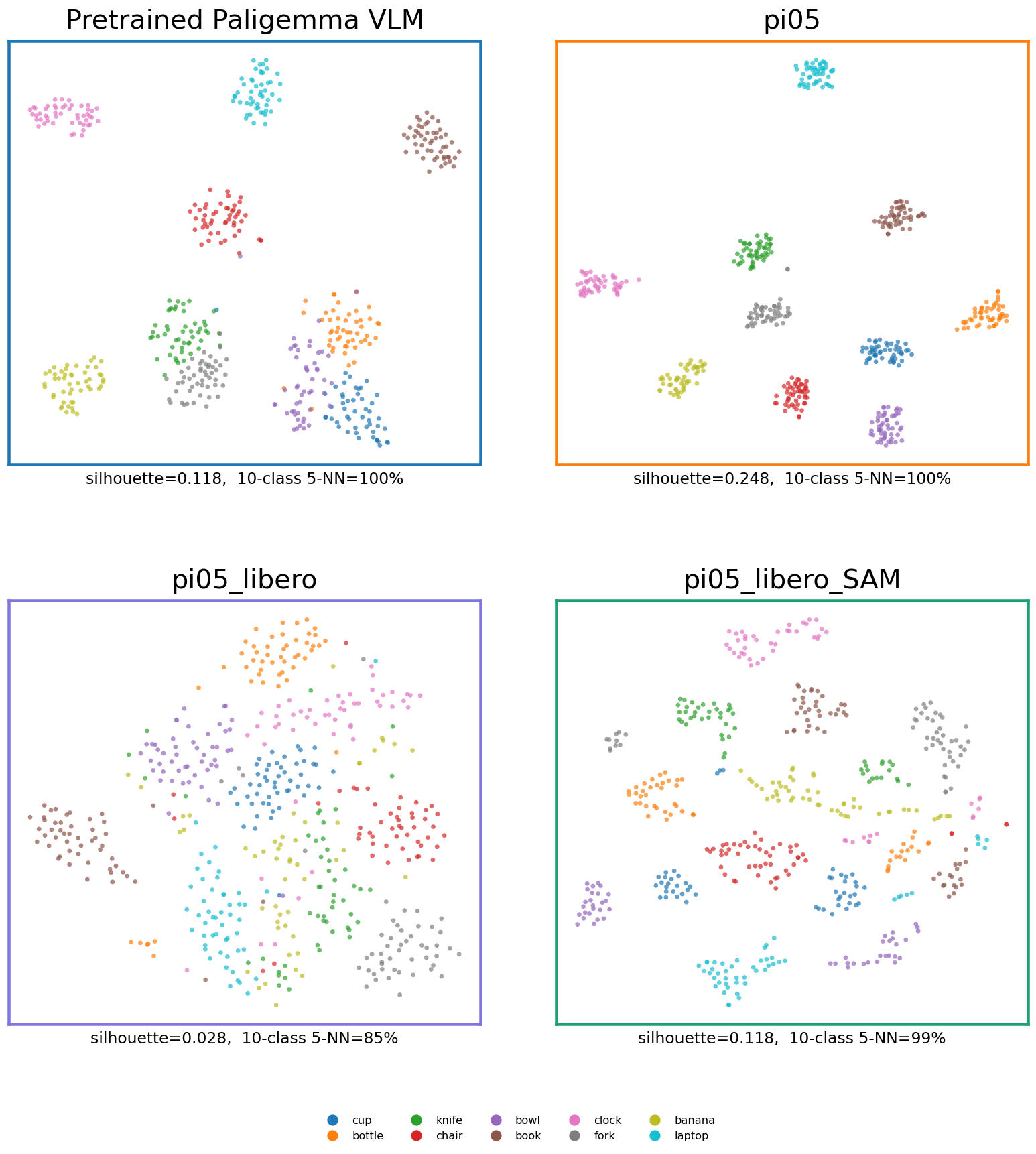}
    \caption{t-SNE plot of object-token representation probe for the Paligemma VLM, \(\pi_{0.5}\)\_base, \(\pi_{0.5}\), and \(\pi_{0.5}\)\_SAM}
    \label{fig:tsne}
    \vspace{-1em}
\end{figure}

We next verify the changes SAM finetuning induces in two ways. First, we conduct a representation-level probe on the vision-language embedding space, inspired by \cite{kachaev2025don}, where we use a probe to analyze object representations across four PaliGemma VLM variants: the base released model 
, the PaliGemma within the \(\pi_{0.5}\)\_base VLA (pretrained on robotics data, but not finetuned on LIBERO), the PaliGemma within the standard finetuned \(\pi_{0.5}\)\_libero checkpoint, and the PaliGemma backbone within the SAM-finetuned \(\pi_{0.5}\)\_libero. We conduct the probe on the COCO dataset \cite{lin2014microsoft} and visualize with a t-SNE plot. As seen in Fig. \ref{fig:tsne}, in the pretrained VLM and even the base VLA model, the object categories form coherent semantic clusters. In contrast, the representations are disrupted after standard finetuning, but then implicitly remedied by finetuning with SAM. 

Additionally, we quantify the curvature of the landscape with the sharpness metric on the full-parameter model and the maximum eigenvalue of the Hessian \(\lambda_{max}\) on the action head. We verify in Table \ref{tab:regularization_splits} that the sharpness value and \(\lambda_{max}\) of the Hessian are both lower when finetuning with SAM, indicating that a flatter loss landscape was achieved while improving instruction-following performance. This, in combination with Table \ref{tab:train_results} also validates that the landscape was optimized to find flatter and wider loss basins without any catastrophic forgetting of training tasks.

\subsection{RQ 3: Component Analysis}
\label{RQ3}

\begin{figure*}[t]
    \centering
    \includegraphics[width=0.9\linewidth]{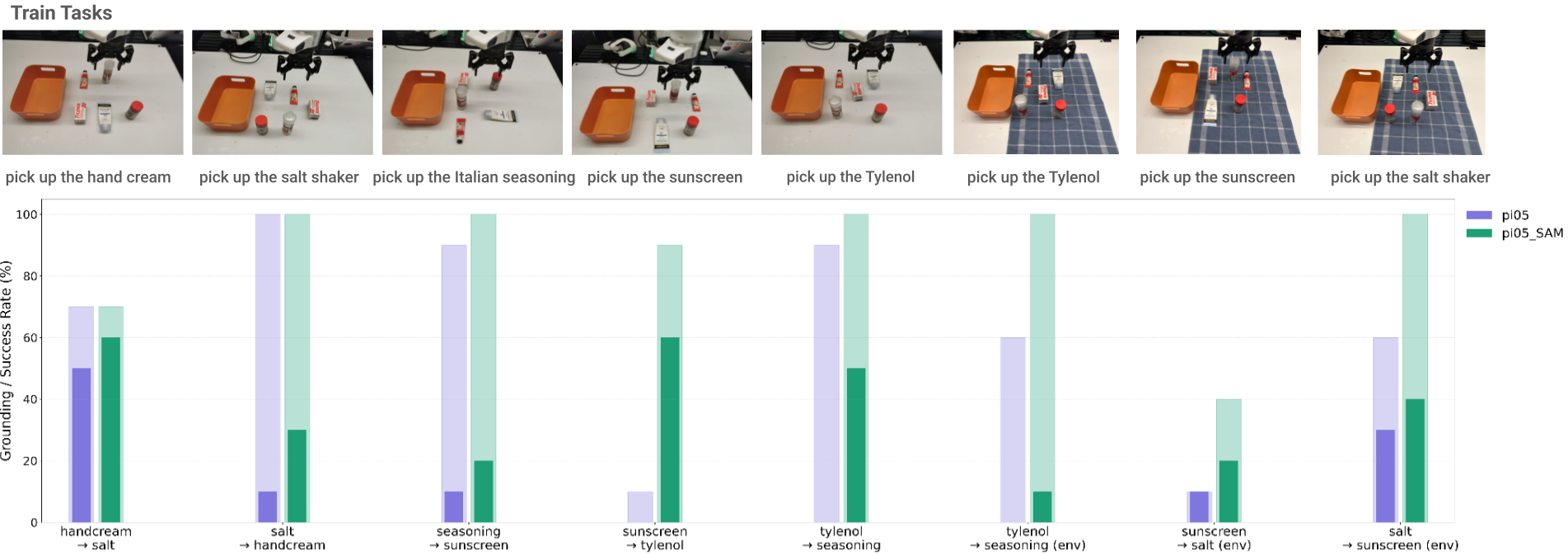}
    \caption{Real-world experiments on five object layouts and a counterfactual instruction at inference time. Each bar shows the grounding rate (translucent) and task success rate (solid).}
    \label{fig:real_world_results}
    \vspace{-1em}
\end{figure*}

\label{ablation}
To better understand how SAM operates on a VLA architecture, we quantify where sharpness manifests in a standard VLA model and whether SAM should be applied globally or selectively to certain components. We first quantify sharpness in each of the vision, language, and action components as well as the full model, by finding the worst-case loss within some fixed radius of perturbed parameters. We observe from the default VLA that the action-head of the model has the highest isolated sharpness of 0.015, around 6 times more than the isolated sharpness in the language or vision components of the model backbone. Thus, with SAM finetuning, the largest isolated reduction in sharpness is also in the action head, where sharpness is reduced to 0.01 while the sharpness of the full model reduces from 0.016 to 0.011. These findings suggest that when trained on limited data with flow-matching loss, the small-parameter action head may be a bottleneck for a well-conditioned gradient signal. However, sharpness of each component in isolation is not indicative of how model components interact as components are coupled.

We then perform a component-wise ablation by separating the \(\pi_{0.5}\) model into it's vision, language, and action components and selectively applying the SAM algorithm to only the parameters there while maintaining the AdamW optimizer elsewhere. 
We find empirically that global application of SAM consistently outperforms component-wise application in task success. The overall task success rate on LIBERO-PRO is improved over the default VLA model when SAM is isolated to each component, but remains around 37\% for all three, which is lower than global SAM. For SAM isolated to the language head, the action head dominates the overall sharpness with 0.022, and similarly, for SAM isolated to the vision encoder, the action head sharpness of 0.018 dominates. Although isolating SAM to the action head reduces the sharpness in the action head in a targeted way, we find that shifting this component towards flatter minima alone is not sufficient for preserving pretrained representations overall.
This indicates that the curvature induced by low-data finetuning is not localized to a single component but propagates through the VLA network. Additionally, we hypothesize that applying the SAM step only in the direction of certain modalities without co-evolving other components results in bottlenecks in other parts of the model. We further verify the utility of global SAM with OpenVLA-OFT, which does not have a separate action head, and observe similar performance gain.
\section{Evaluation in Real World}

To further evaluate the ability of SAM to mitigate instruction blindness in VLAs, we design a set of real-world evaluations using the DROID \cite{khazatsky2024droid} setup for five pick-and-place tasks shown in Fig. \ref{fig:real_world_results}. For each scene, we collect 50 demonstrations for a basket pick-and-place task and induce visual bias by only collecting demonstrations for one task per scene layout. 
Inspired by LIBERO-Object, our tasks consist of specific object layouts of five household objects: salt shaker, seasoning bottle, sunscreen, Tylenol, and a tube of hand cream.
Across the dataset, each object has demonstrations for how to grasp it. We then finetune $\pi_{0.5}$ on this biased multi-task dataset with and without SAM and evaluate with a counterfactual language instruction at test time for each scene. While the objects remain the same, each test task involves picking up the object at a new location. We also introduce environment perturbations where the background is changed for three of the tasks to evaluate robustness of vision-language grounding. 

We verify that finetuning with SAM results in better grounding success and task success for the model while finetuned on the exact same real-world dataset. While the default finetuned model achieves an average 13.8\% success rate, the model finetuned with SAM achieves 36.3\%, reflecting an improvement of 163\%. We observe that the improvement in correct grounding is smaller on some real-world tasks, likely because of the objects being close to the original DROID training distribution. However, even on tasks when the default model can correctly ground the object, it fails to accurately grasp it, reflecting some uncertainty in the action.
\section{Limitations}
While SAM significantly improves instruction following, several limitations remain. First, SAM approximately doubles the compute relative to standard finetuning. While this overhead is modest compared to the cost of additional data collection or pretraining, it does require enough memory. Parameter-efficient alternatives may alleviate this and we leave these methods to future work.
Second, SAM does not explicitly prevent drift by constraining the model around pretrained initializations. Combining SAM with methods like SWA or parameter merging may become best practice.
Finally, while SAM yields substantial improvements in instruction following, absolute success rates still have significant room for improvement. This suggests that sharp loss landscapes are a significant but not sole contributor to instruction blindness and completing OOC tasks.
We view our results as establishing a stronger baseline rather than a complete solution, and expect further gains when combined with complementary approaches.
\section{Conclusion}
\label{sec:conclusion}
We find that preserving flatness while finetuning improves instruction following significantly across three benchmarks and in real-world experiments. Our analysis reveals that this specifically improves object grounding and that global application of SAM works better over selective, despite the highest sharpness occurring in the action head. We do not intent SAM to be a single comprehensive solution for preserving representations, but rather show that it is complementary to existing techniques and can be stacked with other regularization and guidance-based strategies to provide a stronger foundation for instruction-following. We show that explicitly preserving flatness and considering how pretrained neighborhoods shift during model design is crucial, and hope this work encourages the robotics community to consider loss landscape geometry when seeking to develop generalist robot foundation models.


\clearpage

\balance    


\bibliographystyle{IEEEtran}
\bibliography{refs}  

\end{document}